\documentclass[10pt,onecolumn,letterpaper]{article}

\usepackage{times}
\usepackage{epsfig}
\usepackage{graphicx}
\usepackage{amsmath}
\usepackage{amssymb}
\usepackage{array}
\usepackage{adjustbox}
\usepackage{array}
\usepackage{cellspace}
\usepackage{hhline}

\graphicspath{{figures/}}
\usepackage[utf8]{inputenc} 
\usepackage{multirow}
\usepackage{placeins}
\usepackage{lineno}
\usepackage{color}
\usepackage{hyperref}
\usepackage{subcaption}
\pdfoutput=1

\definecolor{greenao}{rgb}{0.0, 0.5, 0.0}

\newcommand*{\mathcolor}{}
\def\mathcolor#1#{\mathcoloraux{#1}}
\newcommand*{\mathcoloraux}[3]{  \protect\leavevmode
	\begingroup
	\color#1{#2}#3  \endgroup
}

\definecolor{111}{rgb}{0,1,0}
\definecolor{00}{rgb}{1.00000000000000000000,.49803921568627450980,0}
\definecolor{222}{rgb}{0.8,0.9568627451,0.56470588235}
\definecolor{111}{rgb}{0.8,0.9568627451,0.56470588235}

\usepackage[table]{xcolor}

\begin{document}
%%%%%%%%% TITLE
\title{Exploiting the ConvLSTM: Human Action Recognition using Raw Depth Video-Based Recurrent Neural Networks}

\author{Adrian Sanchez-Caballero,  David Fuentes-Jimenez, \\
	 Cristina Losada-Gutiérrez\\
Universidad de Alcal\'a\\
{\tt\small \{adrian.sanchez,d.fuentes\}@edu.uah.es}\\
{\tt\small\{cristina.losada\}@uah.es} }
\maketitle

\begin{abstract}
		As in many other different fields, deep learning has become the main approach in most computer vision applications, such as scene understanding, object recognition, computer-human interaction or human action recognition (HAR). Research efforts within HAR have mainly focused on how to efficiently extract and process both spatial and temporal dependencies of video sequences. In this paper, we propose and compare, two neural networks based on the convolutional long short-term memory unit, namely ConvLSTM, with differences in the architecture and the long-term learning strategy. The former uses a video-length adaptive input data generator (\emph{stateless}) whereas the latter explores the \emph{stateful} ability of general recurrent neural networks but applied in the particular case of HAR. This stateful property allows the model to accumulate discriminative patterns from previous frames without compromising computer memory. Experimental results on the large-scale NTU RGB+D dataset show that the proposed models achieve competitive recognition accuracies with lower computational cost compared with state-of-the-art methods and prove that, in the particular case of videos, the rarely-used stateful mode of recurrent neural networks significantly improves the accuracy obtained with the standard mode. The recognition accuracies obtained are 75.26\% (CS) and 75.45\% (CV) for the stateless model, with an average time consumption per video of 0.21 s, and 80.43\% (CS) and 79.91\%(CV) with 0.89 s for the stateful version.
\end{abstract}

\section{Introduction}

Human behavior understanding has received great interest in computer vision researchers in the last decades due to the broad variety of possible applications. Recognizing and understanding human activity is essential in applications such as automated video surveillance, health care services, human-computer interaction, autonomous driving or video analysis.

Human action recognition (HAR) has been historically studied using visual image sequences or RGB videos~\cite{poppe2010survey,guo2014survey,herath2017going} through different approaches, first with handcrafted feature descriptors~\cite{blank2005actions,laptev2008learning,bregonzio2009recognising,sadanand2012} and, later, with deep learning-based techniques~\cite{ji20123d,simonyan2014two,varol2017long,baccouche2011sequential}. Handcrafted feature-based methods perform well on small datasets, whereas in the case of large datasets the performance of deep neural networks (DNN) is better.

Since the rise of affordable real-time depth sensors, many studies have focused on using depth videos for human action recognition. These depth sensors provide images with rich 3D structural information of the scene with benefits compared to RGB videos such as invariance to changes in lighting conditions, textures or colors, reliability for obtaining human silhouette, and preservation of personal-privacy, which generates great interest in some domains as video surveillance. Furthermore, the 3D structure information of depth maps allows estimating the human joints positions~\cite{shotton2011real} also known as the 3D skeleton, which supposes itself a different data modality for HAR and removes the need of using motion capture (MOCAP) systems.

As in the case of RGB-based HAR works, the first depth-based studies employed methods based on handcrafted descriptors~\cite{vieira2012stop,klaser2008spatio,wang2012robust,oreifej2013hon4d}, but eventually works using deep learning techniques also became the main approach. The rise of the deep learning techniques is directly associated with the concurrent appearance of public large-scale datasets in HAR like NTU RGB+D~\cite{shahroudy2016ntu}. DNN-based approaches have been proved~\cite{wang2019comparative} to be more robust and suitable for challenging large datasets than handcrafted features-based methods. 

Among the different DNN and depth-based approaches for HAR, many of them modify the input to generate depth motion maps~\cite{wang2015convnets} or dynamic images~\cite{wang2016large,wang2015action,wang2018depth,xiao2019action,wu2019hierarchical} so as to encode the spatio-temporal information of a complete video into some few images through color and texture patterns. Besides, convolutional neural networks (CNNs), successfully used in image processing tasks, can be extended to a third dimension~\cite{singh2019depth,liu20163d} (3D CNN) to handle the temporal extension of videos. A widely employed alternative to CNNs is the recurrent neural network (RNN)~\cite{shahroudy2016ntu,li2018independently,shi2017learning,du2015hierarchical}, where \emph{neurons} belonging to different time steps of a sequence are interconnected. It allows learning temporal patterns in any type of sequential data, making RNNs adequate for applications such as speech recognition, handwriting recognition, and time series prediction. One particular RNN which solves the exploding or vanishing gradient problem is the long short-term memory (LSTM)~\cite{hochreiter1997long}, so it is able to successfully learn patterns in long sequences like videos by stacking several layers. Regarding the HAR problem, this type of neural network is commonly used after some feature extraction with, for example, a CNN~\cite{nunez2018convolutional, donahue2015long}, or with 3D information of skeletons~\cite{park2016depth, song2017end, shahroudy2016ntu} as input.  One limitation of LSTMs in HAR is they cannot directly learn spatio-temporal features from a sequence of images. This limitation was overcome replacing the Hadamard product of the original LSTM with the convolution operation (ConvLSTM)~\cite{xingjian2015convolutional}, changing vectors for tensors in the hidden states. To the best of our knowledge, there have been no studies where raw videos are directly fed to an RNN or, more specifically, to an LSTM. It is still more unlikely to find a study that uses an LSTM network in \emph{stateful} operation mode for action recognition, where all the potential of this kind of architecture is exploited. 

In this work, we propose and analyze two novel implementations of recurrent neural networks based on ConvLSTMs that receive depth videos as input. The usage of these videos instead of RGB-based sequences allows us to leverage not only the already mentioned technical benefits of depth modality but also its capacity to be used in privacy-preserving video surveillance, because of the absence of texture and color in images that permits face recognition. We train and test our deep learning models on the large scale dataset NTU RGB+D~\cite{shahroudy2016ntu} using only the depth video sequences, which were recorded with Microsoft Kinect 2~\cite{zhang2012microsoft} sensor. Because of the great number of samples that it contains, NTU RGB+D is especially suitable for DNN-based methods. We use raw depth videos as input and feed them to a ConvLSTM network without any prior calculation like skeleton positions, optical flow or dynamic images. As a major contribution, we study and propose a novel implementation of the unusually used \emph{stateful} capability for LSTM layers, in order to fully exploit the long term memory. We also utilize various techniques and training strategies from deep learning theory~\cite{breuel2015effects,smith2018disciplined}. In particular, to reduce the usual subjectivity in the learning rate choice for training, we use a learning rate range test to estimate the optimal values once the batch size is fixed. Furthermore, we use a cyclical learning rate~\cite{smith2017cyclical} to improve convergence and reduce over-fitting to yield competitive results. Additional techniques used in this paper are batch normalization~\cite{ioffe2015batch}, LeakyReLU activations~\cite{maas2013rectifier} and the usage of an average pooling layer instead of the common fully connected layers at the top of the neural network. The latter reduces drastically the number of parameters and improves model generalization. Finally, a video-length-adaptive input data generator has been designed to fully exploit the temporal dimension of long videos.

The purpose of this work is to show that relatively simple and efficient neural networks, like a ConvLSTM, can perform reasonably well in human action recognition tasks when their capabilities are exploited. In particular, the rarely used \emph{stateful} mode of LSTMs is studied. Very competitive results and close to state-of-the-art methods in HAR are obtained. It is also shown that the stateful version outperforms the conventional LSTM operation mode.

The rest of this paper is organized as follows. In section~\ref{sec:related-works}, previous works related to HAR are explained, giving special emphasis to depth-based methods. Section~\ref{sec:rnn-architectures} includes the architecture description of the RNNs. Subsequently, in section~\ref{sec:training-stage}, the training stage of the proposed models is explained. Section~\ref{sec:experimental-results} shows and discusses the obtained experimental results. Finally, the paper is concluded in section~\ref{sec:conclusion}.

\section{Related work}
\label{sec:related-works}
% RGB works
Initial works on HAR used visual images recorded with standard RGB cameras and methods with handcrafted features~\cite{blank2005actions,laptev2008learning,bregonzio2009recognising,sadanand2012}. Motivated by the success in image processing tasks, deep learning methods began to be applied also for videos, typically with architectures such as 3D convolutions (3DCNN) and RNNs~\cite{ji20123d,baccouche2011sequential}. In particular, a very common framework found in the literature is the two-stream neural network~\cite{simonyan2014two,feichtenhofer2016convolutional,wang2016temporal}, where one stream operates on RGB frames whereas the other tries to learn motion using optical flow as input. The optical flow is pre-computed with handcrafted methods, which involves a high computational cost. To alleviate this, N. Crasto \emph{et al.}~\cite{crasto2019mars} proposed using a feature-based loss that mimics the motion stream in the two-stream 3D CNN and removes the need for using optical flow. Most of the deep learning-based works on HAR with RGB videos put the effort into solving the problem of how to treat efficiently the temporal dimension of videos. The previous methods use th third dimension in convolutions to deal with the extra dimension. 

However, the existence of long videos, which is inherent for certain human actions, may not allow the neural network to process discriminative features due to memory limitations, failing to recognize these actions. This long-term problem in 3DCNNs is especially studied in~\cite{varol2017long}, where they propose to increase the temporal size of the input at the cost of reducing spatial resolution, or in~\cite{SUN201833} by building motion maps to represent motion from videos of any length. Another alternative is to use RNNs such as LSTM units to learn temporal patterns from features previously extracted with spatial CNNs. This scheme together with a temporal-wise attention model and a joint optimization module to fuse output features is used in~\cite{wang2018human}. Other researchers have used deep learning to estimate optical flow~\cite{ilg2017flownet} instead of using traditional and computationally expensive methods. Additionally, novel spatial and temporal pyramid modules for CNN are proposed and aggregated to a Spatial-Temporal Pyramid Network (S-TPNet) in~\cite{ZHENG2019446} to learn effective spatio-temporal pyramid representations of videos. 

% RGB+D works
Instead of extracting human pose estimations from RGB images, many other works~\cite{sung2012unstructured,LIU201574,Hu2015CVPR,hsu2016,wang2017scene,kong2017max,liu2019rgbd} combine directly RGB with depth modality like 3D skeleton or depth maps to leverage the advantages from both types of data.

% Depth works
Regarding depth-based works for HAR, there exist different approaches depending on how 3D information is treated. As explicitly mentioned in~\cite{xiao2019action}, depth-based videos are usually divided into three categories depending on the nature of input: human skeleton-based, raw depth-video-based and a combination of both. The evolution and progress of approaches on these three categories have been affected by the growth of deep learning in the last years, especially in computer vision, leading most recent studies to use this technique. In this regard, P. Wang \emph{et al}.~\cite{wang2018rgb} elaborated a very complete survey of recent studies using deep learning in human motion recognition tasks.

In the first category, 3D positions of human body skeletons must be previously extracted from the depth map in each frame or by using MOCAP systems. There are many different approaches to how the 3D skeleton joint positions can be managed as, for instance, computing multiple joint angles~\cite{park2016depth}, extracting discriminative parts for each human action~\cite{HUANG201884} or finding the best viewpoint for recognition as in~\cite{LIU2019}. Skeleton joint positions, or any other data derived from them, are also fed into neural networks in most recent papers, mainly through RNN-based methods~\cite{wang2014learning, park2016depth, song2017end, liu2016spatio, zhang2017view, nunez2018convolutional}, but also with CNNs~\cite{du2015skeleton,ZHU2019109}.

Secondly, raw depth maps are directly used as input to the model. Different descriptors have been proposed as methods for the classification process, as in~\cite{oreifej2013hon4d, yang2014super, lu2014range}. X. Yang \emph{et al.}~\cite{yang2012recognizing} proposed Depth Motion Maps (DMM) to represent depth videos through a pseudo coloring image. Later, DNN-based architectures with DMMs as input~\cite{wang2015action, wang2015convnets} improved prior results. Alternatively to DMMs, in~\cite{wang2017structured,wang2018depth}, suggested using three pairs of images for video representation using bidirectional rank pooling. Y. Xiao \emph{et al.}~\cite{xiao2019action} have recently worked with multi-view dynamic images, reaching state-of-the-art results in the raw depth maps modality.

Finally, some researchers chose to use both 3D skeleton positions and depth maps and reached good results~\cite{rahmani2017learning, shi2017learning}, taking the benefits from both modalities at the cost of an increase in model complexity. Indeed, the combination of these two types of data is more often used with traditional hand-crafted feature algorithms.

A recent comparative review of action recognition methods~\cite{wang2019comparative} asserts that skeletal data-based models have achieved better accuracy and robustness than depth-based ones. Nevertheless, 3D skeleton joints have some known drawbacks: general information loss, potential failures of 3D positions extraction and the impossibility of action detection involving human-object interactions. In addition, 3D skeleton joints can not be directly extracted with a camera, unless a MOCAP system is used, which is not plausible in most applications. On the other hand, depth-based techniques are more similar to how human vision works but with extra 3D information and can be recorded with a camera and immediately used as input to deep learning models without any intermediate calculation. Studies related to this modality are valuable in research fields as computer vision and scene understanding.

\section{RNN Architectures}
\label{sec:rnn-architectures}

As explained before, there can be found many different approaches to human action recognition problems in the literature. Recently, the usage of deep learning has become the main tool in these studies due to its good results. It has been used both in extracting features and decision processes, usually making use of a CNN~\cite{nunez2018convolutional}. It has been shown that temporal pooling techniques used for reducing a video to some representative images yield good performances, but a preprocessing is required. RNNs have also been used although, generally, as part of the decision process, so that temporal dependencies are taken into account, or with high-level representations such as skeleton data.

\begin{figure}[ht]
	\centering
	\includegraphics[width=\linewidth]{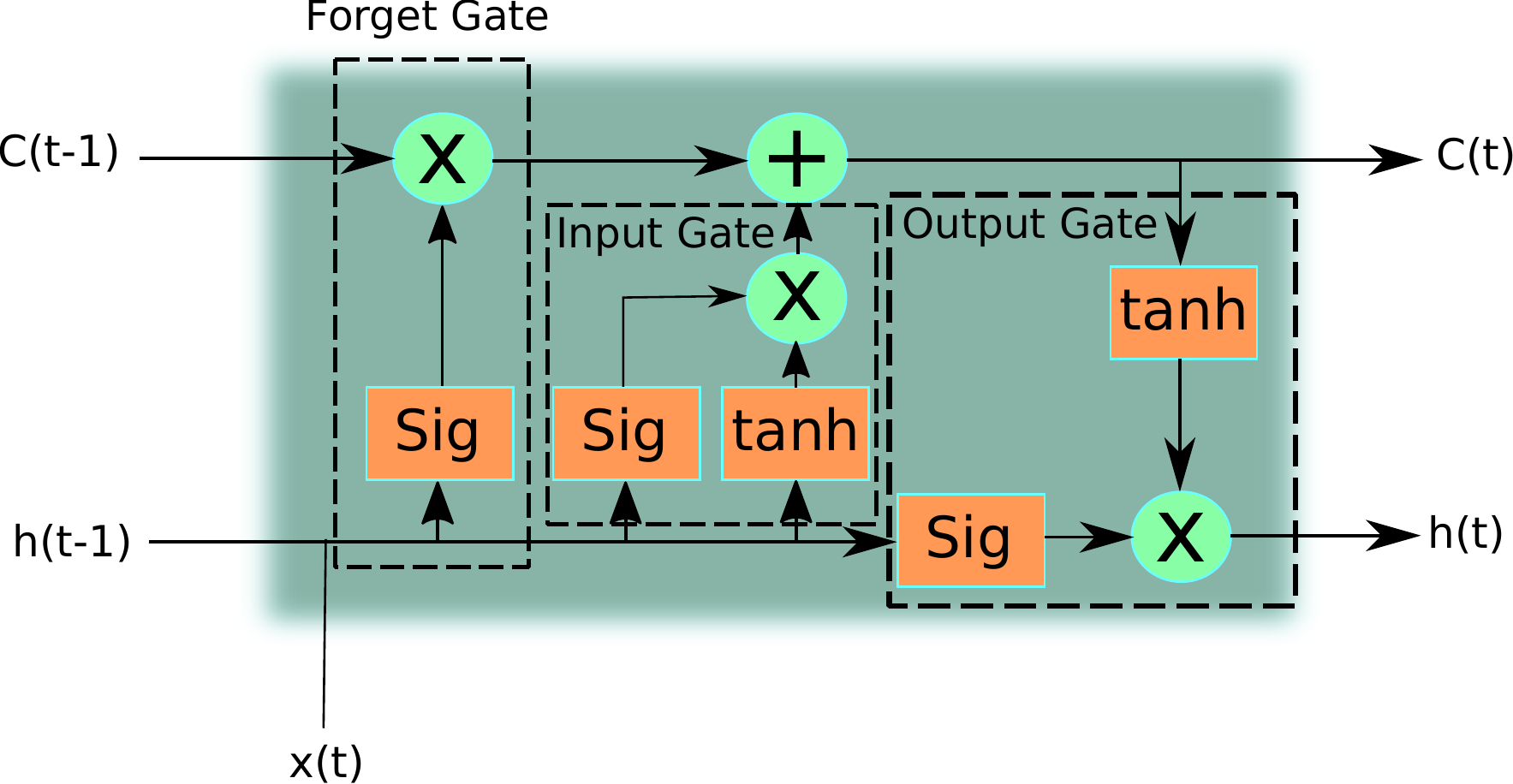}
	\caption{A representation of an LSTM unit, where the cell state vector at time step $t$, $C(t)$, is gradually modified taking into account the previous hidden state $h(t-1)$ and input $x(t)$ first by a forget gate, which determines which values from $C(t-1)$ can be removed, then an input gate, which performs the update of values in the cell state, and finally the output gate, which decides what is going to be output.}
	\label{fig:LSTM}
\end{figure}

Among RNN networks, one of the most popular is the LSTM~\cite{hochreiter1997long}. It is characterized by including a memory cell or cell state, which is modified over time steps through three different gates: input, forget and output (Fig.~\ref{fig:LSTM}). This property allows LSTM networks to model long-term dependencies.

Additionally, in~\cite{xingjian2015convolutional} a modified LSTM layer is proposed, namely the convolutional LSTM or ConvLSTM, and applied it directly to videos for weather forecasting. A ConvLSTM layer possesses a similar mathematical structure as the original LSTM~\cite{hochreiter1997long} but replacing Hadamard product operation with matrix convolutions and using 3D tensors instead of vectors. Consequently, ConvLSTM layers are able to encode both spatial information, as CNNs do, and temporal patterns extracted from previous frames. This makes ConvLSTM a good choice for modeling spatiotemporal sequences like videos, removing any type of the previous encoding or preprocessing.

The short-term memory of an LSTM layer is represented by the cell state $C_{t}$, which in the case of ConvLSTM is a 3D tensor, whereas the long-term memory is reflected in the trainable weights inside the gates. However, short-term memory is actually the novel property LSTMs introduce. Ideally, the cell state will not be reset until the entire time sequence is fed to the network. Thus, the cell state can contain full-sequence information, but in the real world, this situation is not always feasible. Training data are provided to neural networks in batches with sizes that are restricted by the CPU/GPU memory capacity of computers. When the used data consist of videos, it is necessary to find a balance between the number of frames in each input sample (the bigger, the more long-term dependencies our model will extract) and samples in each batch (generally the more, the better the model will generalize and avoid over-fitting), both subjected to the hardware memory limitation. 

The LSTM \emph{stateless} mode resets its cell state after each batch is processed and the weights are updated. Most studies use LSTMs in stateless mode since these layers usually operate on already extracted features or simplified data, so the memory limit is not significant. However, there exists a solution to this limitation through what is called the \emph{stateful} mode of an LSTM. With this mode, the LSTM layers preserve the cell state from the previous batch removing any memory restriction. This property allows LSTMs to handle videos of very different lengths as usually happens with HAR samples, where the information from previous frames can be extremely useful.

These properties are also present in the convolutional version of the LSTM layer, ConvLSTM, and therefore, they can be applied to videos. In this paper, we make a performance comparison between stateless and stateful networks on a challenging depth-based HAR dataset~\cite{shahroudy2016ntu}.

% 	\begin{figure*}[ht]
% 		\centering
% 		\includegraphics[width=\linewidth]{ConvLSTM-graph-v2}
% 		\caption{Schematic of the stateless ConvLSTM network used in this paper.}
% 		\label{fig:stateless-architecture}
% 	\end{figure*}

\subsection{Stateless ConvLSTM network}

LSTM stateless mode is set by default in most machine learning libraries and usually omitted in the papers that use this architecture. This operation mode does not require any particular data preparation (as opposed to stateful mode) and performs well in many cases. 

\begin{figure*}[ht]
	\centering
	\includegraphics[width=\linewidth]{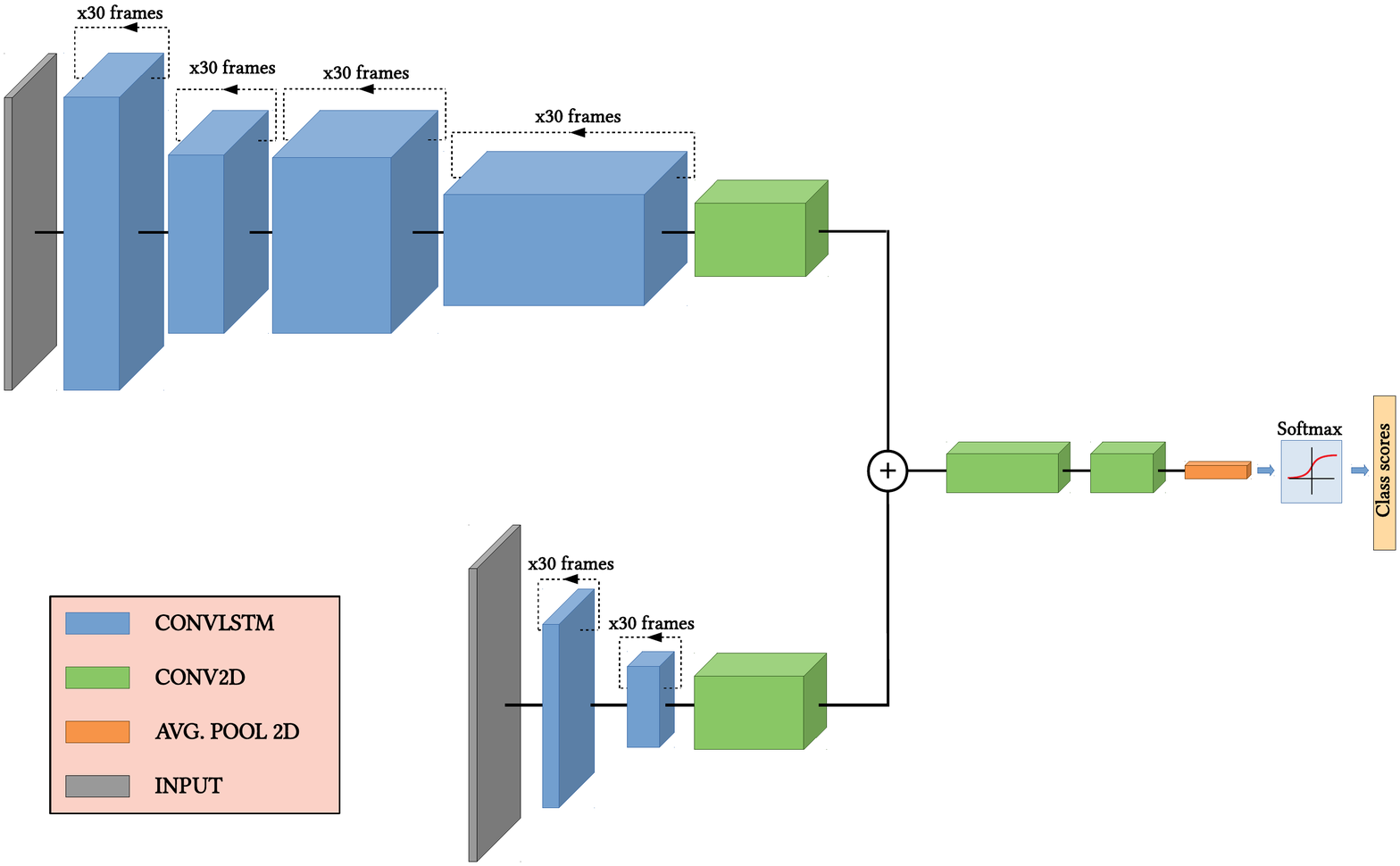}
	\caption{Schematic of the stateless ConvLSTM network used in this paper.}
	\label{fig:stateless-architecture}
\end{figure*}

The stateless ConvLSTM network proposed in this work contains two stages: a recurrent block, which extracts features directly from the video frames, and a decision block with convolutional and pooling layers. Furthermore, the network contains two parallel branches (main branch and support branch) that are fused afterward through addition. Fig.~\ref{fig:stateless-architecture} shows a general block representation of this architecture where both branches can be seen.

The main branch is composed of 4 stacked ConvLSTM layers with Batch Normalization (BN) after each one (see Fig.~\ref{fig:stateless-architecture}). BN reduces internal covariance, contributing to speed up the training. The ConvLSTM layers have 32, 32, 128 and 256 $3\times3$ filters, respectively.  The last recurrent layer removes the temporal dimension and leads to a convolutional layer with 128 $3\times3$ filters. At this point, the support branch is added up to the main one. 

The support branch (lower branch in Fig.~\ref{fig:stateless-architecture}) has fewer layers but bigger filters. The input passes through two ConvLSTM layers with 8 $7\times7$ and 16 $5\times5$ filters, respectively. Next, there is a convolutional layer with 128 $3\times3$ filters before the summation with the main branch. A more complete description can be seen in Table~\ref{tab:stateless-arch} including kernel and strides properties of convolutions. The employed activations are LeakyReLU (\textit{Leaky Rectified Linear Unit}~\cite{maas2013rectifier}) functions. This type of function follows the expression $f(x)=x$ if $x\geq0$ and $f(x)=\alpha x$ if $x<0$, with $\alpha=0.3$, and it has proven to be more efficient~\cite{xu2015empirical} compared with the standard ReLU activation. 

These convolutional layers start the decision block. After the summation, there are other two convolutional layers with 128 $3\times3$ filters that precede 2D global average pooling and \emph{softmax} activation, producing a vector that includes the class likelihoods standardized to the unit.

\begin{table*}
	\centering
	\begin{tabular}{ l || c || c  }
		\hline
		\textbf{Layer} & \textbf{Parameters} & \textbf{Output size} \\
		\hline
		\hline
		\multicolumn{3}{c}{\textbf{Support branch}} \\
		\hline
		\hline
		Input & - & (30, 64, 64, 1) \\
		\hline
		ConvLSTM & k=(7, 7), s=(2, 2) & (30, 29, 29, 8)\\
		\hline
		Batch Normalization & \multicolumn{2}{c}{-} \\
		\hline
		ConvLSTM & k=(5, 5), s=(2, 2) & (1, 13, 13, 16) \\
		\hline
		Batch Normalization & \multicolumn{2}{c}{-} \\
		\hline
		Conv2D & k=(1, 1), s=(2, 2) & (13, 13, 128) \\
		\hline
		\hline
		\multicolumn{3}{c}{\textbf{Main branch}} \\
		\hline
		\hline
		Input & - & (30, 64, 64, 1)\\
		\hline
		ConvLSTM 1 & k=(3, 3), s=(1, 1) & (30, 64, 64, 32) \\
		\hline
		Batch Normalization & \multicolumn{2}{c}{-} \\
		\hline
		Activation & \multicolumn{2}{c}{LeakyReLU} \\
		\hline
		ConvLSTM 2 & k=(3, 3), s=(2, 2) & (30, 31, 31, 32) \\
		\hline
		Batch Normalization & \multicolumn{2}{c}{-} \\
		\hline
		Activation & \multicolumn{2}{c}{LeakyReLU} \\
		\hline
		ConvLSTM 3 & k=(3, 3), s=(1, 1) & (30, 64, 64, 128) \\
		\hline
		Batch Normalization & \multicolumn{2}{c}{-} \\
		\hline
		Activation & \multicolumn{2}{c}{LeakyReLU} \\
		\hline
		ConvLSTM 4 & k=(3, 3), s=(2, 2) & (1, 15, 15, 256) \\
		\hline
		Batch Normalization & \multicolumn{2}{c}{-} \\
		\hline
		Activation & \multicolumn{2}{c}{LeakyReLU} \\
		\hline
		Conv2D & k=(3, 3), s=(1, 1) & (13, 13, 128) \\
		\hline
		\textbf{Add Main Branch + Support Branch} & - & (13, 13, 128) \\
		\hline
		Activation & \multicolumn{2}{c}{LeakyReLU} \\
		\hline
		Conv2D & k=(3, 3), s=(2, 2) & (6, 6, 128) \\
		\hline
		Batch Normalization & \multicolumn{2}{c}{-} \\
		\hline
		Activation & \multicolumn{2}{c}{LeakyReLU} \\
		\hline
		Conv2D & k=(1, 1), s=(1, 1) & (6, 6, 60) \\
		\hline
		Global Average Pooling 2D & - & (1, 1, 60) \\
		\hline
		Activation & \multicolumn{2}{c}{Softmax} \\
		\hline
	\end{tabular}
	\caption{Summary of stateless ConvLSTM network architecture. k: kernel size, s: stride.}
	\label{tab:stateless-arch}
\end{table*}

\subsection{Stateful ConvLSTM network}

The stateful ConvLSTM architecture is slightly simpler than stateless. It consists of a single branch and the structure is very similar to the main branch of the Stateless ConvLSTM network. 

The recurrent block contains 4 ConvLSTM layers with 32, 64, 128 and 256 $3\times3$ filters and BN after each one. After the third ConvLSTM layer, a regular 2D convolution has been placed to reduce the number of features and, consequently, the overall network parameters. The decision block is composed of 2 convolution layers with 128 $3\times3$ filters, accompanied by BN and Leaky ReLU activations as in the stateless architecture. Next, a final convolution reduces the number of features to match the number of classes, which precedes Global average pooling and \emph{softmax} activation. A detailed description of every layer of the stateful model is reported in Table~\ref{tab:stateful-arch}.

The complexity of the stateful network falls essentially on the particular training and data arrangement, which will be explained in the next section. 

\begin{table}[t]
	\centering
	\begin{tabular}{ l || c || c}
		%{ >{\ct}m{0.16\textwidth} | >{\ct}m{0.10\textwidth} | >{\ct}m{0.12\textwidth}  }
		
		\hline
		\textbf{Layer} & \textbf{Parameters} & \textbf{Output Size} \\
		\hline
		\hline
		Input & - & (8, 64, 64, 1)\\
		\hline
		Stateful ConvLSTM 1 & k=(3, 3), s=(1, 1) & (8, 64, 64, 32) \\
		\hline
		Batch Normalization & \multicolumn{2}{c}{-} \\
		\hline
		Stateful ConvLSTM 2 & k=(3, 3), s=(2, 2) & (8, 31, 31, 64) \\
		\hline
		Batch Normalization & \multicolumn{2}{c}{-} \\
		\hline
		Stateful ConvLSTM 3 & k=(3, 3), s=(1, 1) & (8, 31, 31, 128) \\
		\hline
		Batch Normalization & \multicolumn{2}{c}{-} \\
		\hline
		Stateful ConvLSTM 4 & k=(3, 3), s=(2, 2) & (1, 15, 15, 256) \\
		\hline
		Batch Normalization & \multicolumn{2}{c}{-} \\
		\hline
		Conv2D 1 & k=(3, 3), s=(2, 2) & (7, 7, 128) \\
		\hline
		Batch Normalization & \multicolumn{2}{c}{-} \\
		\hline
		Activation & \multicolumn{2}{c}{LeakyReLU} \\
		\hline
		Conv2D 2 & k=(3, 3), s=(2, 2) & (3, 3, 128) \\
		\hline
		Batch Normalization & \multicolumn{2}{c}{-} \\
		\hline
		Activation & \multicolumn{2}{c}{LeakyReLU} \\
		\hline	
		Conv2D 3 & k=(1, 1), s=(1, 1) & (3, 3, 60) \\
		\hline
		Global Average Pooling 2D & - & (1, 1, 60) \\
		\hline
		Activation & \multicolumn{2}{c}{Softmax} \\
		%Add (Main Branch+Shortcut)& - & (Width/2, Height/2, a)\\
		\hline
	\end{tabular}
	\caption{Summary of stateful ConvLSTM network architecture. k: kernel size, s: stride.}
	\label{tab:stateful-arch}
\end{table}

\section{Training stage}
\label{sec:training-stage}

We use the NTU RGB+D dataset~\cite{shahroudy2016ntu}, which is one of the largest human action datasets that include videos in RGB, depth-maps, 3D-skeletons and infrared. It contains 56880 samples with one or more subjects performing a particular action. Videos have been recorded using three simultaneous \textit{Microsoft Kinect II}~\cite{zhang2012microsoft} sensors and, thus, providing multi-view scenes. Resolution of RGB videos is $1920 \times 1080$ pixels, whereas for depth-map and infrared videos it is $512 \times 424$ pixels. 3D skeletal data provide three-dimensional locations of 25 main human body joints for every frame. The database contains 60 human actions within three well-defined groups: daily actions, medical conditions, and mutual actions.

\begin{figure}[ht]
	\centering
	\includegraphics[width=\linewidth]{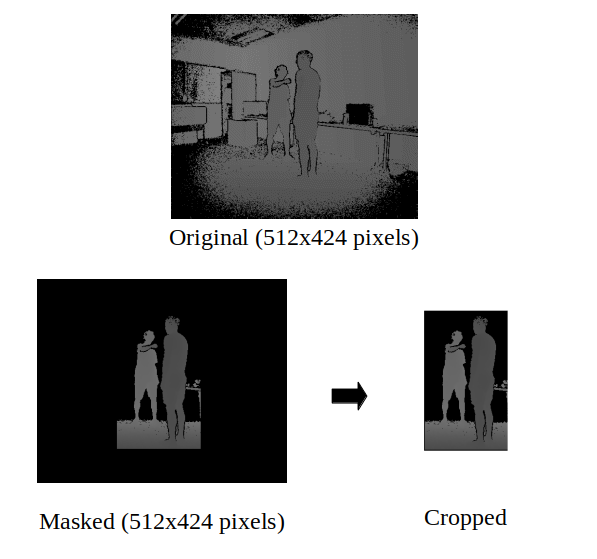}
	\caption{Example of one original video frame (up) from the NTU RGB+D dataset~\cite{shahroudy2016ntu} and the cropping process of its masked version (down).}
	\label{fig:sample-crop}
\end{figure}

This work only makes use of the depth-map modalities and adapts the two data evaluations suggested in~\cite{shahroudy2016ntu} by which the training and test are divided: cross-subject (CS) and cross-view (CV). The provided images are foreground masked versions to improve the compression ratio of files and alleviate the processes of downloading and managing such a big amount of data. They are then cropped to the movement area of the action, as shown in Fig.~\ref{fig:sample-crop}. Finally, the model itself takes the cropped images and re-scales them to $64 \times 64$ pixels to build the network input. Below, it is explained how data is organized and fed as input to the network, which is the main difference between stateless and stateful mode.

\subsection{Training of the stateless ConvLSTM network}

Data arrangement concerns how training and test samples are generated from data and fed to the model, and usually has a big influence not only on the ability to train a neural network but also on the final accuracy of the model. It is required a good understanding of the network architecture and taking into account the dataset properties to get an optimum data arrangement. 

\begin{figure}[]
	\centering
	\includegraphics[width=.75\linewidth]{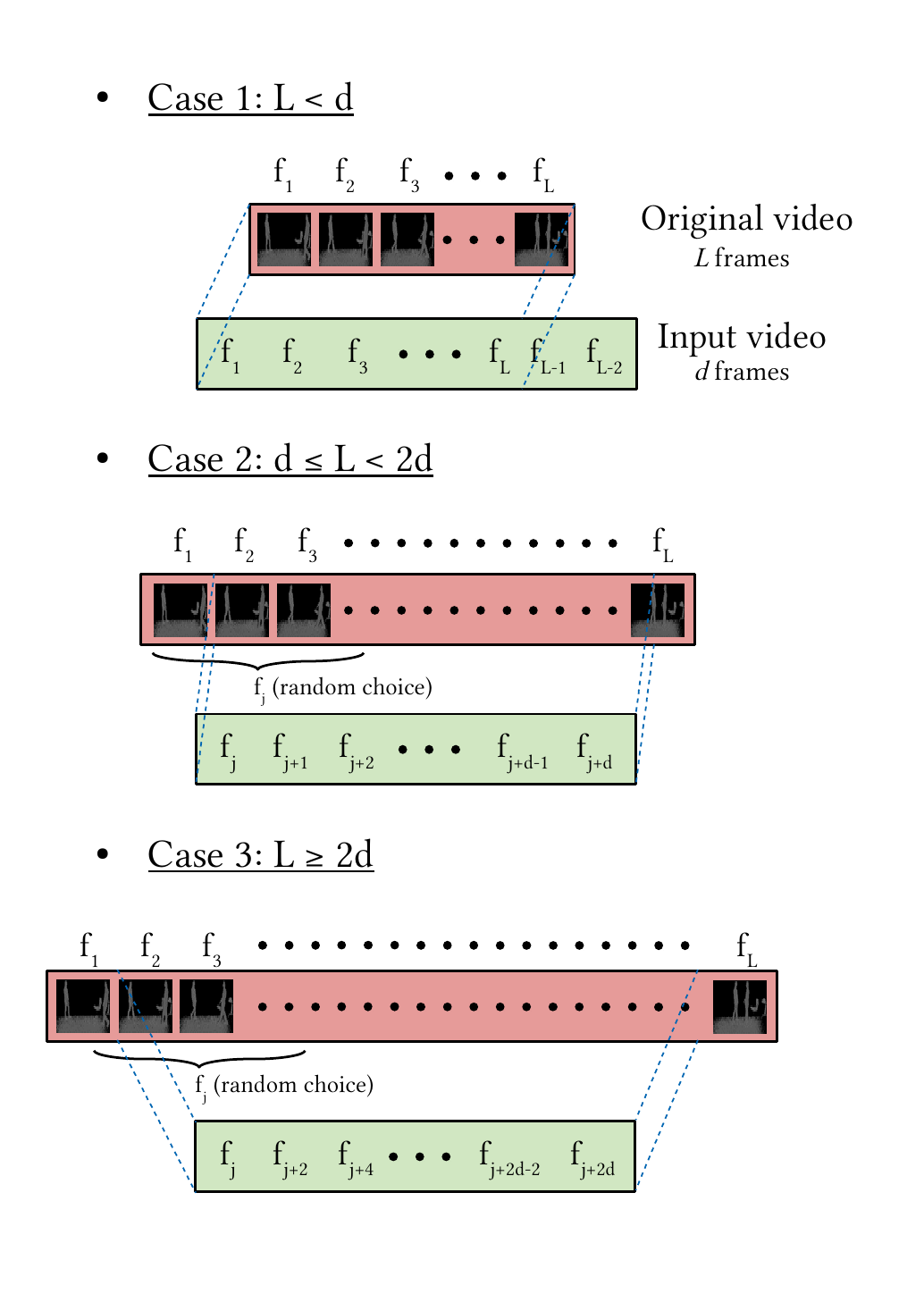}
	\caption{Window selection scheme performed by the stateless input data generator. Let $L$ be the number of frames of a video from the dataset, the input data generator select $d$ frames to build the input video that will be fed to the neural network. When $L<d$, the original video is extended by repeating the last frames until reaching $d$ frames. In the second case, an initial frame $f_{j}$ is randomly chosen provided that the $d$ subsequent frames fit inside the original video. Finally, when $L \geq d$, the initial randomly chosen frame $f_{j}$ is followed by $\{f_{j+2},f_{j+4},...,f_{j+2d}\}$ to cover a region of size $2d$ in the original video.}
	\label{fig:window-selection}
\end{figure}

The temporal sequences that are fed to our network are 30 frames long, which corresponds to 1 second of a video. This value has been experimentally chosen following the previously mentioned balance between the number of frames in the input sample and batch size, which is set to 12, but also subject to the hardware memory limitation. 
However, in the NTU dataset, video lengths go from 26 to 300 frames. There are only a few videos shorter than 30 frames. In this case, some of the final frames have been smoothly repeated until the desired length is reached. When videos are longer, the starting point of the 30-frames temporal window is randomly selected and, in case of very long videos, it also skips frames uniformly in order to cover a wider video range (see Fig.~\ref{fig:window-selection} for an explanatory illustration). These strategies have proven to achieve a better performance of the network for action recognition without increasing the number of input frames. 

\begin{figure}[]
	\centering
	\includegraphics[width=\linewidth]{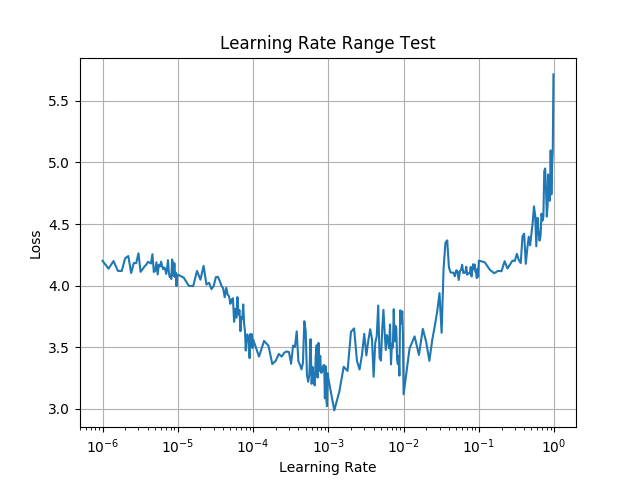}
	\caption{A learning rate range test performed for the stateless ConvLSTM on NTU RGB+D dataset (CV evaluation) with a batch size of 12. The interval of values where the loss function decreases define the optimal range for the learning rate. In this figure, it would be between $10^{-5}$ and nearly $10^{-3}$.}
	\label{fig:LR-range-test}
\end{figure}

The training method has been as follows. First, a learning rate range test is performed to find the optimum interval of values, as suggested in~\cite{smith2017cyclical} when using a cyclical learning rate schedule. As mentioned in~\cite{breuel2015effects}, there exists a  dependency between batch size and learning rate, so we have first set the batch size to 12 and then perform the learning rate range test (see Fig.~\ref{fig:LR-range-test}). From the results of this range test, we choose a customized cyclical schedule, which can improve accuracy with faster convergence. In the first 21 epochs, the learning rate moves linearly between a minimum value of $8\times10^{-5}$ and a maximum of $9.8\times10^{-4}$. After that, boundaries are reduced to $10^{-5}$ and $10^{-4}$, respectively. Finally, after epoch 44 the learning rate is fixed to the minimum $10^{-5}$ until training completes 48 epochs. The algorithm \emph{Adam}~\cite{Kingma2014} has been used as optimizer. This algorithm performs a stochastic gradient descent with an adaptive learning rate computed from estimations of first and second moments of the gradients, and it has proven to achieve fast convergence and be computationally efficient with large models and datasets.

Due to the big training times, it has been used a checkpoint technique in training that continuously saves the model weights when validation accuracy improves. Thus, we take the best model between the former 48 epochs and extend training on 27 more epochs using an initial learning rate of $2\times10^{-4}$ that is reduced by half after 4 epochs without accuracy improving.

\begin{figure}[t]
	\centering
	\includegraphics[width=\linewidth]{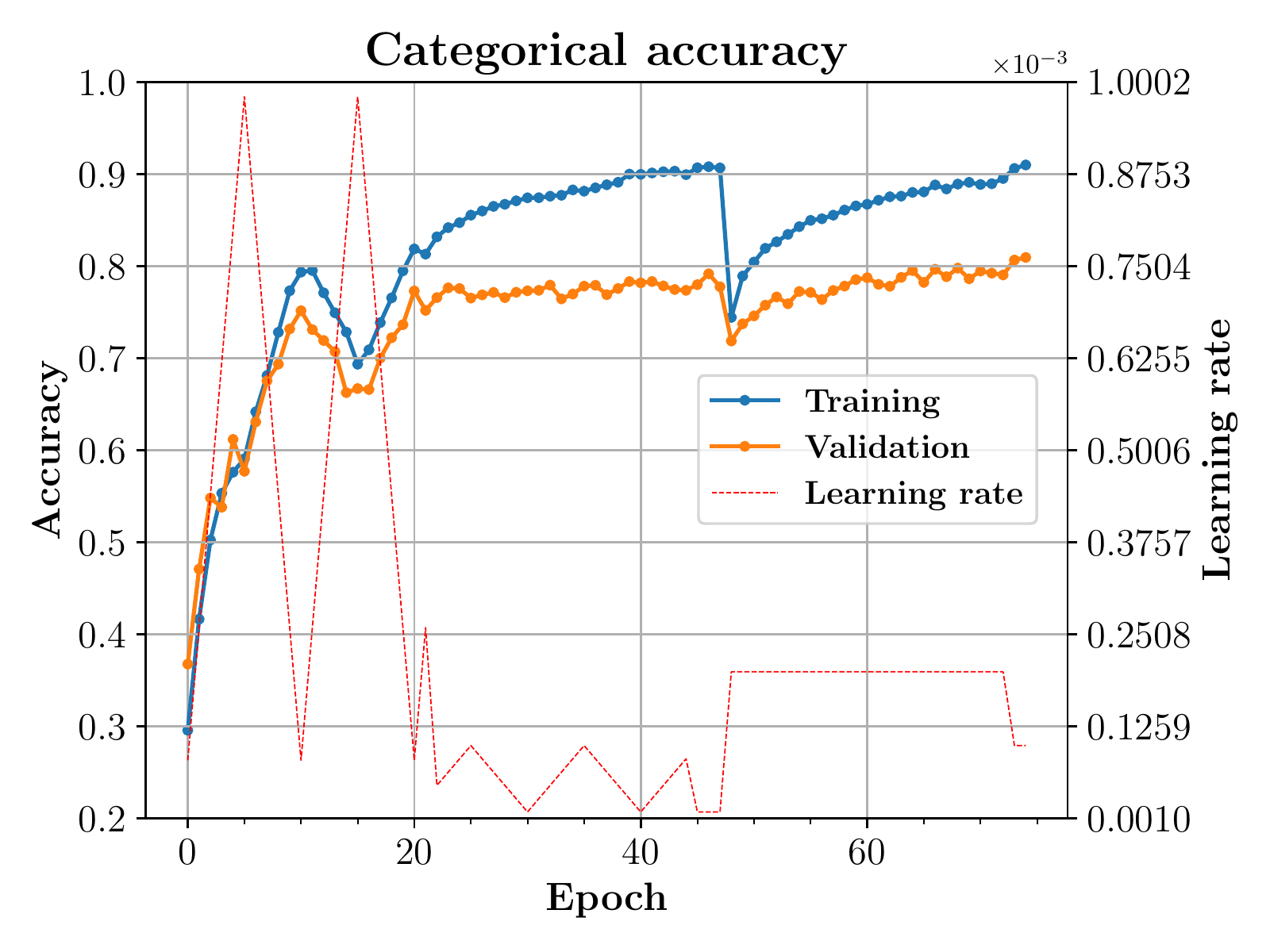}
	\caption{Curves of training and validation accuracies for the stateless ConvLSTM network. In addition, the learning rate schedule used along the epochs is shown.}
	\label{fig:acc-stateless}
\end{figure}

\begin{figure}[t]
	\centering
	\includegraphics[width=\linewidth]{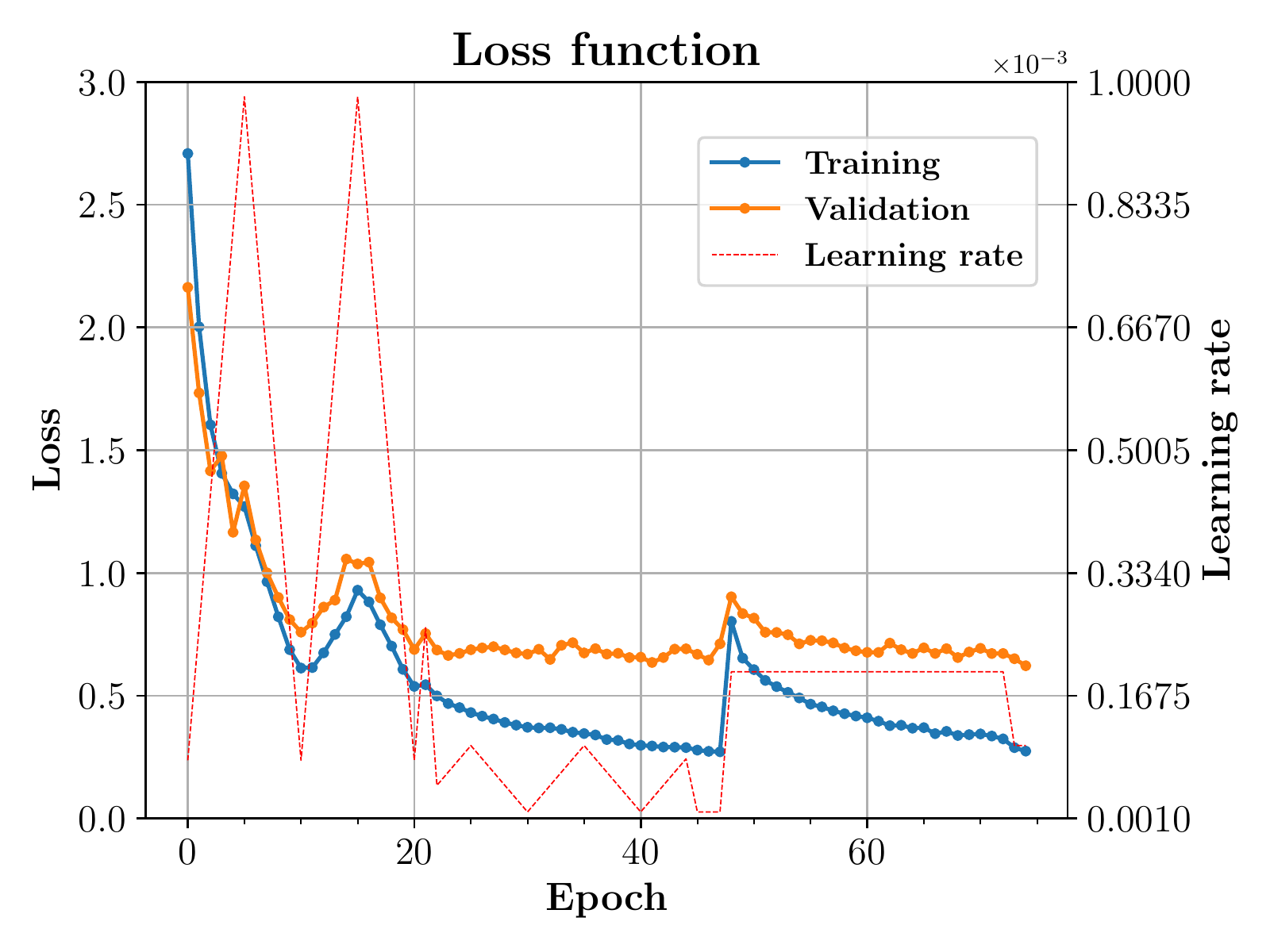}
	\caption{Curves of training and validation loss function for the stateless ConvLSTM network. In addition, the learning rate schedule used along the epochs is shown.}
	\label{fig:loss-stateless}
\end{figure}

The followed learning rate schedule together with the training and validation curves are shown in Fig.~\ref{fig:acc-stateless} and~\ref{fig:loss-stateless} for recognition accuracy and loss function, respectively. Here it can be seen how the variation of learning rate affects accuracy and loss function curves. For instance, the big step at epoch 48 shown in both accuracy and loss function appears due to a significant change in learning rate meaning to find a better minimum of the loss function and to reduce over-fitting.

\subsection{Training of the stateful ConvLSTM network}

Training the ConvLSTM in stateful operation mode requires data preparation as videos have to be sorted by their lengths. This is necessary for the neural network to know where an action ends, and at this point, \emph{reset states} so a new state starts for the next sequences.

\begin{figure}[]
	\centering
	\includegraphics[width=\linewidth]{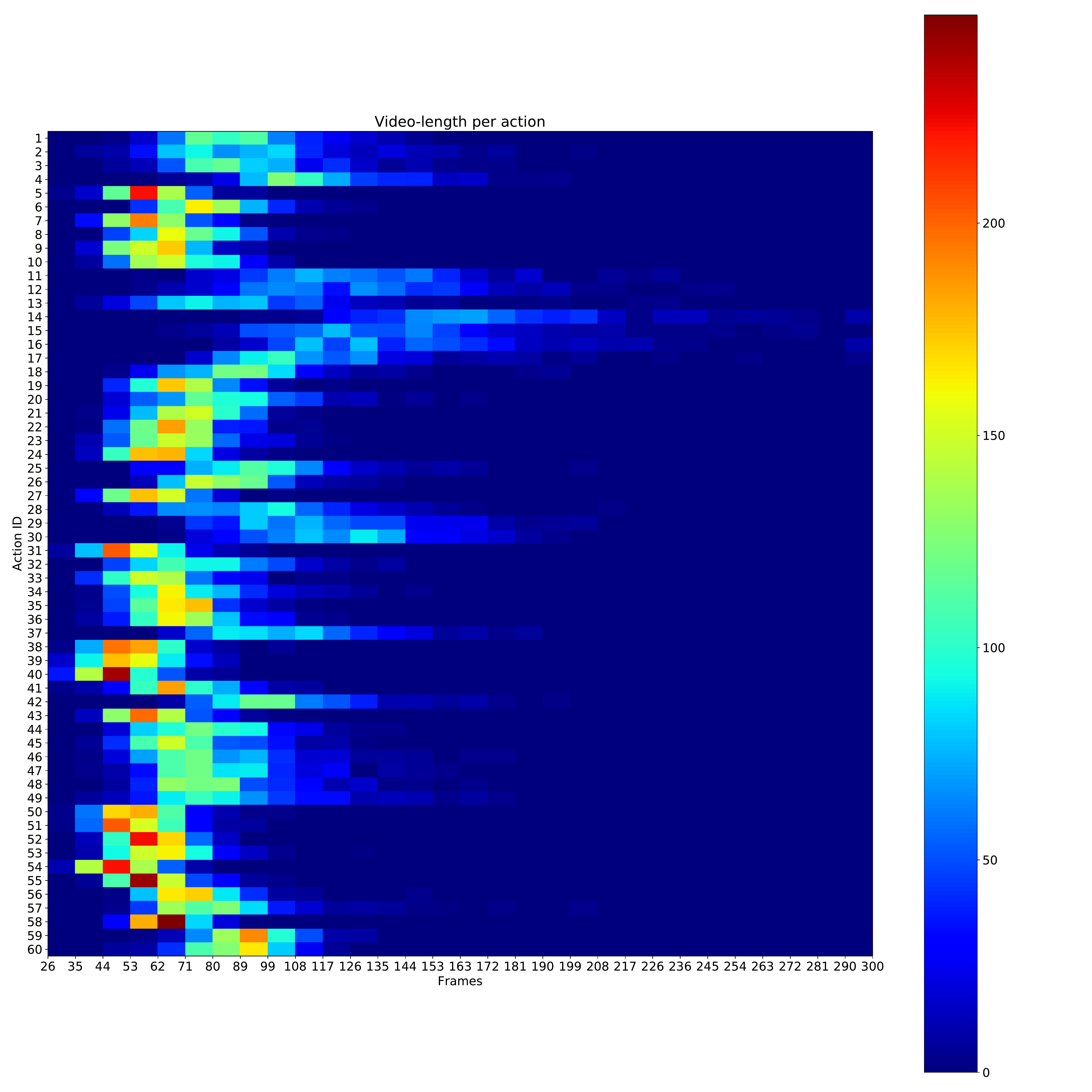}
	\caption{An example of a 2D histogram for video-length per class distribution for the NTU depth database training data (CS evaluation), with automatic bin edges. Note the maximum value of 300 frames corresponds to videos in actions 14, 16 and 17.}
	\label{fig:2d-hist}
\end{figure}

\begin{figure}[!h]
	\centering
	\includegraphics[width=0.75\linewidth]{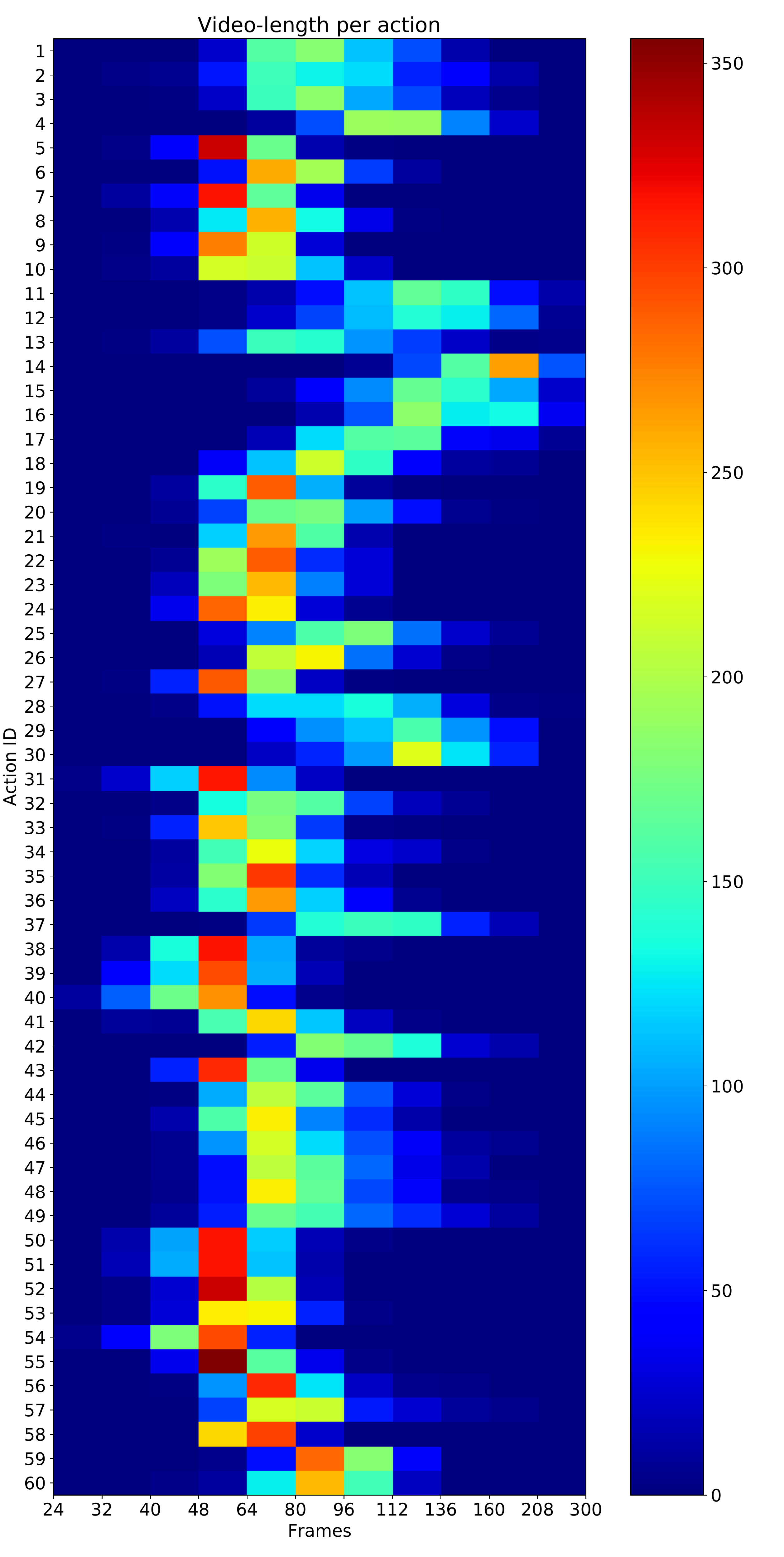}
	\caption{The 2D histogram for video-length distribution per class in NTU depth database with customized bin edges.}
	\label{fig:segmented-2d-hist}
\end{figure}

Therefore, a video-length analysis of the dataset is required to ensure data balance for training. A distribution of the video-lengths in NTU RGB+D dataset is shown in Fig.~\ref{fig:2d-hist} for CS evaluation. It shows big differences of video lengths, which range from 26 to 300 frames, but most of them fall into the 44-90 frames region. Thus, we selected a customized set of bin edges in order to get a slightly more uniform distribution, which can be seen in Fig.~\ref{fig:segmented-2d-hist}. 

The left limit of every bin in Fig.~\ref{fig:segmented-2d-hist} is chosen to be the length of the videos inside that bin. For example, a video of 46 frames is reduced to 40 frames and one of 300 to 208. These discrete lengths are chosen to be multiple of 8, which is set as the number of frames in each temporal window or unit clip that is fed to the neural network at each step. Thus, videos inside the bin of 112 frames will have 14 pieces of 8 frames, i.e. the neural network has to look through 14 different windows until the 112 frames are reached.

Every time the network processes one of these windows, it is able to update weights. If we let the network do this with every window of a video, validation metrics will behave abnormally and a strong over-fitting will appear. To solve this, we make the network process the first half of the video without making weight updates, but preserving the cell states, and then train on the windows that belong to the second half using here the information gained from the previous frames. 

Similarly, we have calculated both training and testing metrics considering that late predictions (final windows) are more reliable than the initial ones, where the state does not contain enough information yet. Therefore, for each video, a weighted average is performed using per-window predictions. The distribution of weights $w(t)$ follows the expression shown in Equation~\ref{eq:1}.

\begin{equation}
w(t) = Nt^{a}
\label{eq:1}
\end{equation} 

\noindent where $t$ is the window number and $N$, a normalization constant. We chose $a=3$, whereas the value of $N$ is video-dependent and adopts the expression $T^{-a}$, where $T$ is the total number of windows in a sequence.

We found that training of the stateful network is more sensitive to learning rate changes than the stateless mode. Therefore, to obtain a non-divergent validation loss, we experimentally found some valid learning rate values. Due to the unusualness of this training, the learning rate range test is not used here. The batch size has been set to 6. As in stateless mode, we used Adam as optimizer and a $25\%$-rate dropout right before the decision block to reduce over-fitting.

\begin{figure}[th]
	\centering
	\includegraphics[width=\linewidth]{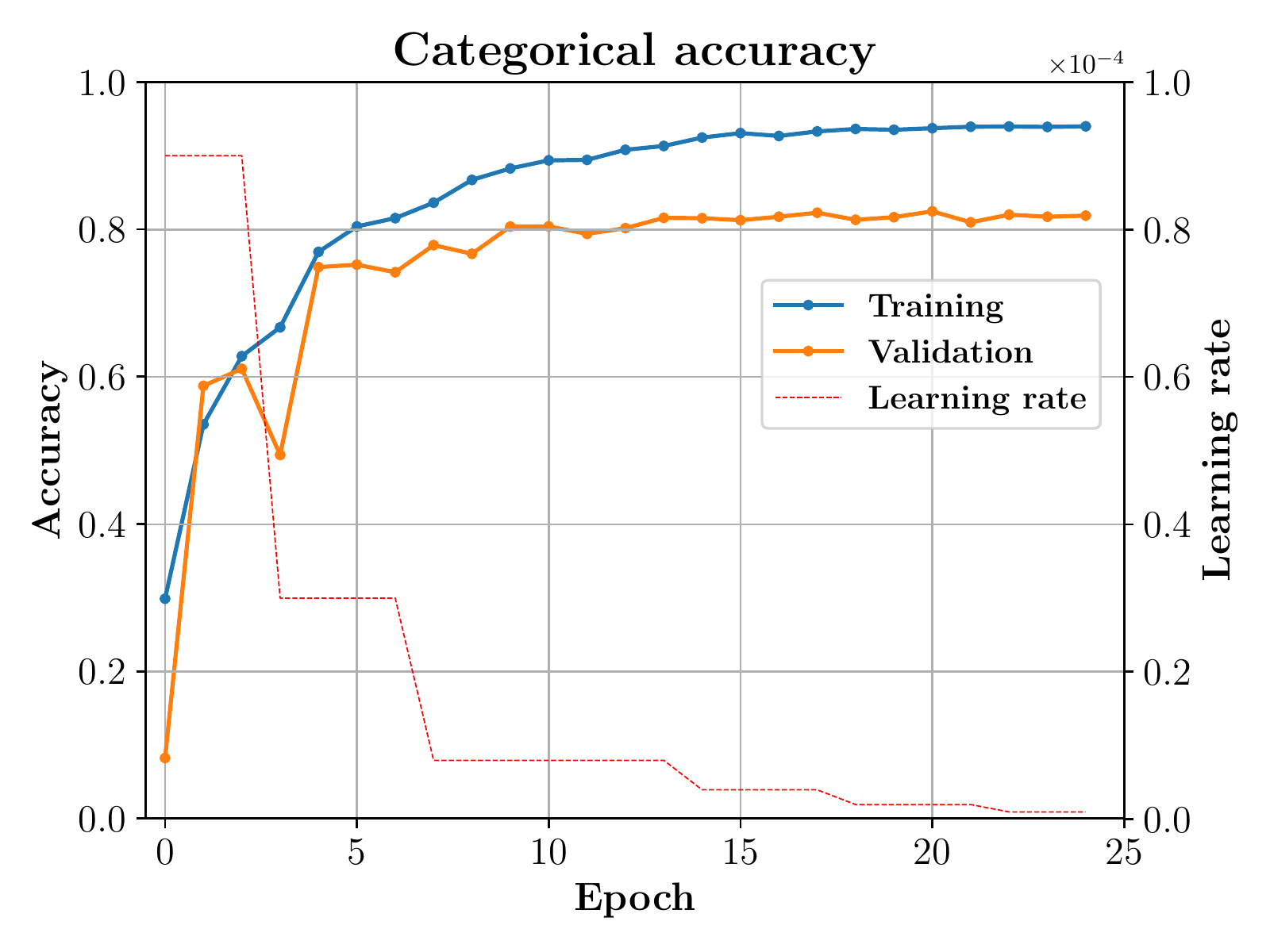}
	\caption{Curves of training and validation accuracies for the stateful ConvLSTM network. In addition, the learning rate schedule used along the epochs is shown.}
	\label{fig:acc-stateful}
\end{figure}

\begin{figure}[th]
	\centering
	\includegraphics[width=\linewidth]{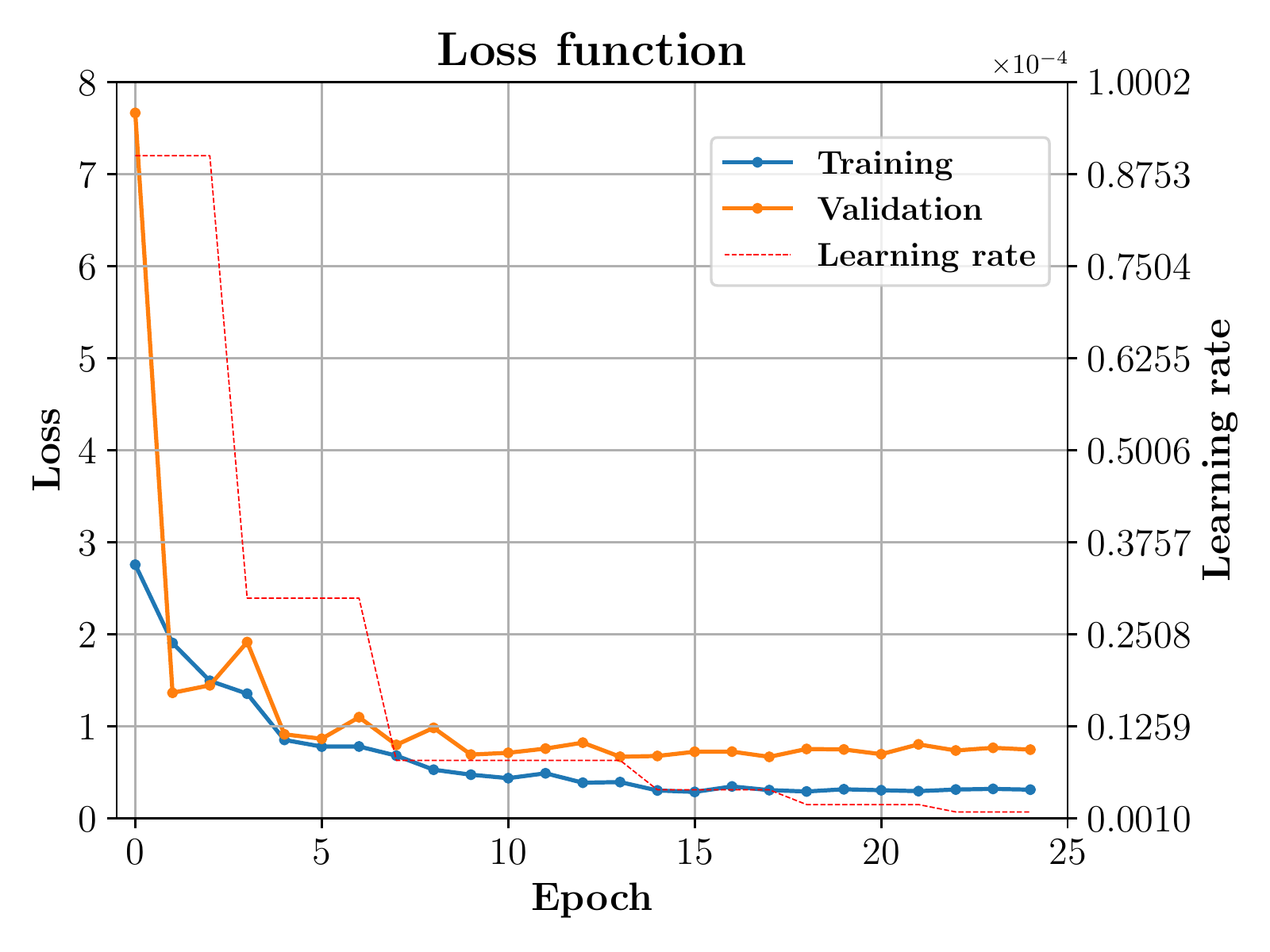}
	\caption{Curves of training and validation loss functions for the stateful ConvLSTM network. In addition, the learning rate schedule used along the epochs is shown.}
	\label{fig:loss-stateful}
\end{figure}

We have experimentally observed that small learning rate values are needed to minimize model divergence in training. The applied learning rate schedule for stateful training has been as follows. The initial learning rate is set to $9\times10^{-5}$, then diminished to $3\times10^{-5}$ in epoch 4, to $8\times10^{-6}$ in epoch 8, to $4\times10^{-6}$ in epoch 15 and, from here, divided by 2 every 4 epochs until complete a total of 25 epochs. The learning rate schedule can be seen in Fig.~\ref{fig:acc-stateful} together with the recognition accuracy curves of training and validation or in Fig.~\ref{fig:loss-stateful}, where curves of training and validation loss functions are also shown. 
In the previous figures, it can be seen that the stateful ConvLSTM network reaches convergence faster than the stateless mode, but with higher computational time per epoch. It is also noteworthy that a relatively small initial value like $9\times10^{-5}$ for the learning rate still causes a big divergence in the validation curve at first epoch (see Fig.~\ref{fig:loss-stateful}). This erratic behavior has been observed in different curves of training at some precise epochs, proving that the stateful operation of the neural network is especially sensitive to the learning rate size. 

\section{Experimental results}
\label{sec:experimental-results}

The NTU RGB+D~\cite{shahroudy2016ntu} dataset has also been used for the test phase of the proposed methods. The authors of this dataset suggest two different evaluations: cross-subject (CS), where 40\,320 samples recorded with 20 subjects are dedicated for training and 16\,560 samples with 20 different subjects for test; and cross-view (CV), where 37\,920 videos were recorded with 2 cameras from different viewpoints and 18\,960 videos from a third different viewpoint for test. 

Results of both models are presented and compared, first, with a discussion of the per-class recognition accuracies and, second, through a comparison with other state-of-the-art methods. It is also included a computational cost analysis. 

\subsection{Recognition performance analysis}

\begin{table*}[t]
	\centering
	\begin{tabular}{ c || l r || l r  }
		\hline
		\bfseries Method & \multicolumn{2}{c||}{\bfseries Top 10 recognized actions} & \multicolumn{2}{c}{\bfseries Top 10 confused actions$^{*}$} \\
		\hline
		\hline
		
		& 1) Falling down & ($95.65\%$) & 1) Put on a shoe $\rightarrow$ Take off a shoe & ($40.22\%$)  \\
		& 2) Hugging & ($94.93\%$) & 2) Reading $\rightarrow$ Writing & ($46.37\%$)  \\
		& 3) Jump up & ($93.84\%$) & 3) Writing $\rightarrow$ Play with phone/tablet & ($46.74\%$)  \\
		& 4) Shake head & ($92.75\%$) & 4) Take off a shoe $\rightarrow$ Put on a shoe & ($51.81\%$)  \\
		\bfseries Stateless & 5) Walking towards & ($92.39\%$) & 5) Sneeze/cough $\rightarrow$ Chest pain & ($52.17\%$)  \\
		\bfseries ConvLSTM & 6) Put on a jacket & ($91.67\%$) & 6) Take off glasses $\rightarrow$ Put on glasses & ($52.54\%$)  \\
		& 7) Salute & ($91.67\%$) & 7) Put on glasses $\rightarrow$ Take off glasses & ($52.90\%$) \\
		& 8) Pushing & ($91.67\%$) & 8) Play with phone/tablet $\rightarrow$ Writing & ($54.71\%$) \\
		& 9) Pick up & ($90.22\%$) & 9) Rub two hands $\rightarrow$ Clapping & ($55.43\%$) \\
		& 10) Kicking & ($88.77\%$) & 10) Eat meal $\rightarrow$ Brush teeth & ($57.61\%$) \\
		& \multicolumn{2}{r||}{\bfseries Average accuracy = $\textbf{92.36\%}$} & \multicolumn{2}{r}{\bfseries Average accuracy = $\textbf{51.05\%}$} \\
		\hline
		& 1) Jump up & ($98.19\%$) & 1) Writing $\rightarrow$ Play with phone/tablet & ($39.13\%$)  \\
		& 2) Walking towards & ($98.19\%$) & 2) Put on a shoe $\rightarrow$ Take off a shoe & ($48.55\%$)  \\
		& 3) Stand up & ($97.83\%$) & 3) Headache $\rightarrow$ Put on glasses & ($50.00\%$)  \\
		& 4) Walking apart & ($97.83\%$) & 4) Play with phone/tablet $\rightarrow$ Writing & ($52.17\%$)  \\
		\bfseries Stateful & 5) Hugging & ($97.10\%$) & 5) Reading $\rightarrow$ Writing & ($52.54\%$)  \\
		\bfseries ConvLSTM & 6) Sit down & ($96.01\%$) & 6) Sneeze/cough $\rightarrow$ Chest pain & ($54.71\%$)  \\
		& 7) Hopping & ($96.01\%$) & 7) Point to something $\rightarrow$ Taking a selfie & ($63.41\%$) \\
		& 8) Falling down & ($95.29\%$) & 8) Clapping $\rightarrow$ Rub two hands & ($65.58\%$) \\
		& 9) Take off jacket & ($94.93\%$) & 9) Back pain $\rightarrow$ Chest pain & ($68.48\%$) \\
		& 10) Put on a hat/cap & ($93.48\%$) & 10) Take off a shoe $\rightarrow$ Put on a shoe & ($69.20\%$) \\
		& \multicolumn{2}{r||}{\bfseries Average accuracy = $\textbf{96.49\%}$} & \multicolumn{2}{r}{\bfseries Average accuracy = $\textbf{56.38\%}$} \\
		\hline
	\end{tabular}
	
	\begin{flushleft}$^{*}$Numbers between parenthesis are the recognition accuracy of true action (before the arrow). \end{flushleft}
	\caption{Top 10 accurate actions and confused pairs for the proposed model, including accuracy recognition per action (CS evaluation).}
	\label{tab:top-10-actions}
\end{table*}

Table~\ref{tab:top-10-actions} shows the top 10 recognized actions together with the 10 worst classified actions for both stateless and stateful networks. Also, the average accuracy of these top 10 classes is given for easier comparison. 

Regarding the stateless ConvLSTM model, almost all the top 10 recognized actions present an accuracy higher than a $90\%$. On the other hand, this model finds difficulties to classify actions like \emph{put on a shoe}, confused with \emph{take off a shoe}, or \emph{reading}, with \emph{writing}, among others. Some common aspect of these classes is that short motions and small objects are part of the discriminatory patterns and they are difficult to process by the model.

Regarding the stateful network, it overcomes the stateless version both within the top 10 recognized and top 10 confused, with some minor exceptions like classes \emph{writing} or \emph{headache}, which slightly decrease their accuracy percentage. On the whole, the top 10 confused actions improve their recognition rate in almost 5\% and the top 10 recognized in more than 4\% compared with the stateless version.

The total average accuracy after testing both models on the NTU RGB+D dataset is $75.26\%$ (CS) and $75.45\%$ (CV) for the stateless ConvLSTM network and $80.43\%$ (CS) and $79.91\%$ (CV) for the stateful ConvLSTM network. This proves that, although it is rarely used in the literature, stateful mode of the conventional LSTM is able to improve dramatically its performance on challenging datasets like NTU RGB+D. 

In the next section, we compare the obtained results and computational costs with state-of-the-art methods.

\subsection{Comparison with state-of-the-art methods}

\begin{table}
	\centering
	\begin{tabular}{c||cc}
		\hline
		\bfseries Method & \bfseries CS & \bfseries CV \\
		\hline\hline
		\bfseries Modality: 3D Skeleton & & \\
		ST-LSTM + Trust Gate (2016)~\cite{shahroudy2016ntu} & 69.2 & 77.7 \\
		Clips + CNN + MTLN (2017)~\cite{ke2017new} & 79.57 & 84.83 \\
		AGC-LSTM (2019)~\cite{si2019attention} & 89.2 & 95.0 \\
		\hline\hline
		\bfseries Modality: Depth & & \\
		Unsupervised ConvLSTM (2017)~\cite{luo2017unsupervised} & 66.2 & - \\
		Dynamic images (HRP) (2018)~\cite{wang2018depth} & 87.08 & 84.22 \\
		HDDPDI (2019)~\cite{wu2019hierarchical} & 82.43 & 87.56 \\
		Multi-view dynamic images (2019)~\cite{xiao2019action} & 84.6 & 87.3 \\
		\hline
		Stateless ConvLSTM  & 75.26 & 75.45 \\
		Stateful ConvLSTM & 80.43 & 79.91 \\
		\hline
		
	\end{tabular}
	\caption{Comparison of total average accuracy (\%) with the NTU RGB+D dataset using different modalities.}
	\label{tab:comparison-results}
	
\end{table}

A performance comparison of the proposed models (stateless and stateful ConvLSTM) with previous state-of-the-arts methods is shown in Table~\ref{tab:comparison-results}. Thanks to the innovative deep learning techniques applied, the models proposed in this paper achieve competitive recognition accuracies on the NTU RGB+D dataset, and overcome other ConvLSTM-based methods like in~\cite{luo2017unsupervised}. Even so, methods that use dynamic images, as in~\cite{wang2018depth,wu2019hierarchical,xiao2019action},  get the highest accuracies on this dataset with depth modality. 

The usage of dynamic images prevents these methods from being used in real-time applications like video-surveillance, health-care services, video analysis or human-computer interaction, because of the high computational cost related to dynamic image generation. To illustrate this, a summary of published average processing times per video from different works is shown in Table~\ref{tab:comparison-times}, accompanied with the results from the two proposed methods. The first four results in the table were estimated in~\cite{wang2018depth} on a different dataset than NTU RGB+D, and using a different hardware. Still, these results are useful to give a general idea of the computational cost of video-based recognition methods. The time consumption of the multi-view dynamic images-based method was computed in~\cite{xiao2019action} using an Intel(R) Xeon(R) E5-2630 V3 CPU running at 2.4 GHz and an NVIDIA GeForce GTX 1080 with 8 GB on videos from the NTU RGB+D dataset. Using the same GPU in the present work, we estimated the average time consumption from 10\,000 random video samples of the same dataset, giving as a result 0.21 s for the stateless ConvLSTM and 0.89 s for the stateful ConvLSTM. Although the time consumption of the stateful version is small and allows real-time application,  it is still slower than the stateless one since the stateful model analyze the whole video regardless of its length. Furthermore, both models are drastically smaller than dynamic images-based methods. Although there is an improvement of around 7\% in accuracy when using these methods, they are approximately around 100 times slower than the methods proposed in this study.

\begin{table}
	\centering
	\begin{tabular}{c||c}
		\hline
		\bfseries Method & \bfseries Time consumption (s) \\
		\hline \hline
		MSFK+DeepID~\cite{wan2015explore} & 41.00 \\
		SFAM~\cite{wang2017scene} & 6.33 \\
		WHDMM~\cite{wang2015action} & 0.62 \\
		DDI+DDNI+DDMNI~\cite{wang2018depth} & 62.03 \\
		Multi-view dynamic images+CNN~\cite{xiao2019action} & 51.02 \\
		\hline
		Stateless ConvLSTM & \bfseries 0.21 \\
		Stateful ConvLSTM & 0.89 \\
		\hline
	\end{tabular}
	\caption{Time consumption comparison of some action recognition methods with available data.}
	\label{tab:comparison-times}
\end{table}

\section{Conclusion}
\label{sec:conclusion}
In contrast to most previous deep learning-based methods in human action recognition, this paper presents two models based on long short-term memory (LSTM) units for the stage of feature extraction from raw depth videos, followed by an ensemble of convolution and average pooling layers for the classification process. Both proposed models use a variant of LSTM, namely ConvLSTM, that leverages the convolution operation to extract spatial and temporal features from a sequence of images. In addition, to exploit the performance of these models several techniques from deep learning theory have been used, such as learning rate range test, cyclical learning schedule or batch normalization. The major contribution of this work is the implementation of two novel schemes to alleviate the memory limitation that appears when working with video sequences. On the one hand, we proposed an input data generator that takes into account the video lengths and allows the neural network to learn long-term characteristics (stateless ConvLSTM). On the other hand, we leveraged the stateful capability of LSTMs (and ConvLSTMs), by which the states of recurrent layers steadily learn along the video preserving spatio-temporal information of previous frames. That is, we assure that the stateful model processes nearly the whole video length. The main advantage of this approach is that, unlike state-of-the-art methods that generate static video representations such as depth motion maps or dynamic images, the proposed end-to-end trainable stateful model can effectively recognize actions belonging to very long and complex videos. Experiment results on the challenging NTU RGB+D dataset show that both proposed models (stateless and stateful ConvLSTM) reach competitive accuracy rates with very low computational cost compared with state-of-the-art methods because of the absence of any preprocessing. Furthermore, it is observed that the stateful ConvLSTM achieves better accuracy rates than standard or stateless ConvLSTM, proving the effectiveness of this uncommon methodology for videos.

The proved success of the stateful mode operation for HAR may open future research lines that integrate this capability to more complex or robust neural networks that improves accuracy rates in some problematic actions. Additionally, one may leverage its very long-term spatio-temporal pattern learning to design models for real-life continuous/online action recognition, with great interest in the video-surveillance field.

\section*{Acknowledgment}
Portions of the research in this paper used the ``NTU RGB+D (or NTU RGB+D 120) Action Recognition Dataset" made available by the ROSE Lab at the Nanyang Technological University, Singapore.

This work has been supported by the Spanish Ministry of Economy and Competitiveness under projects HEIMDAL-UAH (TIN2016-75982-C2-1-R) and ARTEMISA (TIN2016-80939-R) and by the University of Alcalá under projects ACERCA (CCG2018/EXP-029) and ACUFANO (CCG19/IA-024).

\bibliographystyle{ieee}
\bibliography{Convlstm}

\end{document}